\documentclass[11pt]{article}
\usepackage{coling2020}
\usepackage{times}
\usepackage{url}
\usepackage{latexsym}
\usepackage{graphicx}

\colingfinalcopy 

\title{Blending Search and Discovery: Tag-Based Query Refinement with Contextual Reinforcement Learning}

\author{Bingqing Yu \\
  Coveo \\
  Montreal, CA \\
  {\tt cyu2@coveo.com} \\\And
  Jacopo Tagliabue \\
  Coveo Labs \\
  New York, US\\
  {\tt jtagliabue@coveo.com} \\}

\date{}

\begin{document}
\maketitle
\begin{abstract}
We tackle tag-based query refinement as a mobile-friendly alternative to standard facet search. We approach the inference challenge with reinforcement learning, and propose a deep contextual bandit that can be efficiently scaled in a multi-tenant SaaS scenario.
\end{abstract}

\section{Introduction}
\label{sec:intro}

%
%
\blfootnote{
    %
    %
    %
    %
    %
    %
    \hspace{-0.65cm}  
    This work is licensed under a Creative Commons 
    Attribution 4.0 International License.
    License details:
    \url{http://creativecommons.org/licenses/by/4.0/}.
}

\begin{figure}[h]
  \includegraphics[width=\textwidth]{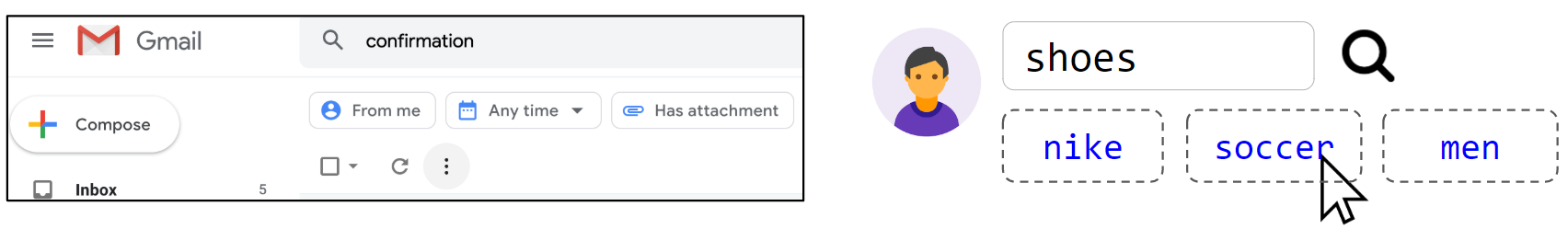}
  \caption{ {\bf Left}: ``Tags'' in \textit{Gmail} search interface.  {\bf Right}: The proposed discovery system as applied to a generic eCommerce shop; catalog, language, data may vary across target shops.}
  \label{fig:google}
\end{figure}

Up to 60\% of the first queries issued by shoppers are one-word queries, and 80\% two words or less\footnote{Data sampled from a network of \textit{500} clients from \textit{Coveo}, a multi-tenant Saas provider in North America.}: in modern digital shops, this often means that shoppers are confronted with tens of thousands of results, and facets listing hundreds of filtering options.\footnote{As an example, at the time of drafting this paper, fashion European behemoth Zalando has 11k results and \textit{hundreds} of items in the \textit{brand} facet for the query ``sandals".} With generic queries, standard facets provide a less  than ideal shopping experience, especially in a world where 60\% of traffic and 50\% of conversions comes from mobile devices~\cite{marketwatch}. ``Tags'' -- also referred to as~\textit{discovery system} (\textit{DS}) hereinafter -- have been proposed (Fig.~\ref{fig:google}, \textit{left}) for smart query refinement: the user is presented directly with one facet value (e.g. \textit{From me}) to refine her intention. We present preliminary findings in building a scalable~\textit{DS} for eCommerce with reinforcement learning (\textit{RL}), robust enough to \textit{scale} in a multi-tenant Saas scenario, across dozens of shops differing in catalog, traffic and language. In contrast to standard query reformulation~\cite{10.1145/3331184.3331362,10.1007/978-3-540-30192-9_58}, we propose a novel refinement experience powered by context-aware~\textit{RL}~\cite{bietti2018contextual,Yurochkin2019OnlineSL}, where the system adjusts itself through constant user feedback; to prepare for online deployment, we also propose an offline evaluation that leverages search logs for reliable performance estimation.

Consider the shopper in Fig.~\ref{fig:google} (\textit{right}), browsing some soccer-related products when he types ``shoes'' in the search bar. In the proposed implementation, the search engine returns a list of results \textit{and} discovery tags; by clicking on, say, ``soccer'', the query will be refined to ``soccer shoes'', and a second search will be issued. In \textit{this} work, we propose an end-to-end \textit{DS} that is able to pick an ``optimal'' tag value (e.g. \textit{Nike}), given a tag type (e.g. \textit{brand})\footnote{For brevity, we sidestep the task of tag type selection \textit{given} a query, as it relates to \textit{parsing} more than \textit{discovery}. From a practical standpoint, please note that most one-word queries are not ambiguous and their compatible tag types can be determined with fairly simple algorithms.}; our focus will be on discovery in a sport apparel scenario, picking values for four types: \textit{sport}, \textit{gender}, \textit{brand} and \textit{price}\footnote{8 sports, e.g. ``soccer"; 2 genders, e.g. ``women"; 349 brands, e.g. ``Nike", and 4 prices, e.g. ``100\$-500\$", in our dataset.}. It is worth mentioning that those four types are selected since they are shared among the majority of our clients in the same vertical; however, nothing crucial hinges on this choice, as eCommerce catalogs are already semi-structured, so that extracting more/different tags is typically no harder than preparing facets for use cases in Information Retrieval.

\section{Experiments}
\label{sec:methods}

\subsection{Dataset}
Data is provided by a partnering digital shop, a mid-sized eCommerce  (annual revenues between 25 and 100 million USD) with traffic and catalog complexity that is most commonly observed in our network, allowing us to assess how well our solution generalizes: given the specificity of our business model -- i.e. providing NLP as a service to hundreds of shops through APIs -- our focus is not so much on scaling \textit{DS} \textit{vertically} to one billion dollar website, but instead scaling \textit{horizontally}, that is providing reliable performances to many mid-size shops with a modest amount of users and data. The training data contains one month of anonymized user sessions, with 227K query sessions and 25K products; each session contains products browsed before the query, which serve as the \textit{context} for our proposed contextual model, and the query itself: products clicked after the query provide ground truth tag values, allowing for model update and validation. The testing dataset is constructed with 50K sessions from a disjoint period. 

\subsection{Models And Results}
We benchmark methods of increasing complexity for~\textit{DS}. One baseline is a popularity model ({\bf POP}), suggesting the most frequently clicked value given a query; {\bf MAB} is a context-independent multi-armed bandit~\cite{Raj2017TamingNB}: a discounted bandit is created for each distinct query in training data, and optimized while picking an arm maximizing reward (in our case, tag value): both models fallback to a global distribution for unseen testing queries. To exploit differentiable representations of the shopping session and the query, we propose a multi-armed deep contextual bandit ({\bf MCM}), taking as input a context vector -- via~\textit{prod2vec}~\cite{10.1145/3383313.3411477,tagliabue-etal-2020-grow} --, concatenated with a query vector -- via BERT~\cite{devlin-etal-2019-bert}\footnote{An alternative setting is training a multilayer perceptron for \textit{each tag value}, and picking the one with the maximum sigmoid output. Preliminary experiments did not show improvement over {\bf MCM}, so only the latter deep method is reported.}. {\bf MAB} and {\bf POP} are trained online, while {\bf MCM} is retrained every 5000 new feedback samples; for all methods, an output matching the attribute of the clicked products is treated as positive feedback; otherwise, negative. Different selection strategies are tested: sampling based on model distribution as an \textit{explorative} strategy; and $\epsilon$-greedy with $\epsilon=0.1$ as a more \textit{exploitative} one. 
\begin{figure}[htp]
    \centering
    \includegraphics[width=14.1cm,height=\textheight,keepaspectratio]{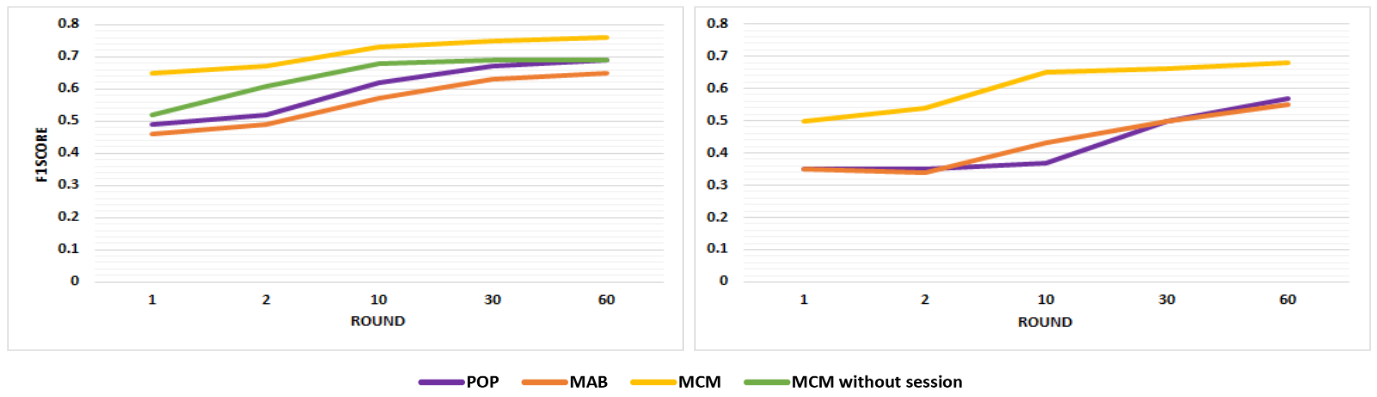}
\caption{Weighted $F1$ score on testing data with explorative (\textbf{left)} and exploitative strategies (\textbf{right}).}
    \label{fig:results}
\end{figure}

Time-ordered training data is split into 60 equal-sized ``rounds"; we report results for the tag type \textit{sport} in terms of weighted $F1$ scores in Fig.~\ref{fig:results}\footnote{Results with the other tag types lead to the same conclusions, thus they are omitted for brevity.}. Our best model combines \textit{explorative} strategy, \textit{context-awareness} and \textit{neural} inference: {\bf MCM} vastly outperforms the baselines, with more than 33\% boost in $F1$ at $round=1$ , and ends with more than 10\% increase in $F1$ at $round=60$\footnote{Using \textit{regret} as measure of success gave mixed results, possibly because of the difference between online and batch updates: we look forward investigating this more in future studies.}; ablation studies show a drop in performance (9\% decrease in final $F1$) with {\bf MCM} trained without session information, proving user context is crucial for personalizing suggestions.

\section{Conclusions}
\label{sec:conclusions}
Narrowing down the result set for generic queries is crucial both to facilitate decision making~\cite{10.1086/651235} and to \textit{nudge} shoppers into exploring the digital shop effectively. In \textit{this} work, we introduced a scalable system to provide personalized discovery tags, and showed how to solve the inference challenge with a context-aware deep~\textit{RL} approach. As next step, we look forward to better understanding \textit{regret} dynamics in our models, and finally  embracing the sequential nature of query refinement, expanding our setting to a multi-step decision process.



\bibliographystyle{coling}
\bibliography{coling2020}

\end{document}